\documentclass[conference]{IEEEtran}
\IEEEoverridecommandlockouts
\usepackage [english]{babel}
\usepackage [autostyle, english = american]{csquotes}
\MakeOuterQuote{"}
\usepackage{cite}
\usepackage{amsmath,amssymb,amsfonts}
\usepackage{algorithmic}
\usepackage{graphicx}
\usepackage{textcomp}
\PassOptionsToPackage{table,xcdraw}{xcolor}  % Pass options BEFORE loading packages
\usepackage{xcolor}                          % Load only once
\usepackage{lipsum}                          % Optional
\usepackage{subfig}
\usepackage{url}

\usepackage{tikz}
\usepackage{colortbl}
\usepackage{float}
\usepackage{hhline}
\usepackage{amsmath}
\usepackage{graphicx}
\usepackage{multirow}
\usepackage{subcaption}  % For caption customization
\usepackage{wrapfig}
\usepackage{multirow}
\usepackage{arydshln} 
\usepackage{multirow}
\usepackage{array}
\usepackage{tabularx}
\usepackage{placeins}

\def\BibTeX{{\rm B\kern-.05em{\sc i\kern-.025em b}\kern-.08em
    T\kern-.1667em\lower.7ex\hbox{E}\kern-.125emX}}
    
\begin{document}

\title{Lab-AI: Using Retrieval Augmentation to Enhance Language Models for Personalized Lab Test Interpretation in Clinical Medicine\\
}

\author{\IEEEauthorblockN{Xiaoyu Wang}
\IEEEauthorblockA{\textit{Department of Statistics} \\
\textit{Florida State University}\\
Tallahassee, FL, USA \\
xw22e@fsu.edu}
\and
\IEEEauthorblockN{Haoyong Ouyang}
\IEEEauthorblockA{\textit{Department of Statistics} \\
\textit{Florida State University}\\
Tallahassee, FL, USA \\
ho23a@fsu.edu}
\and
\IEEEauthorblockN{Balu Bhasuran}
\IEEEauthorblockA{\textit{School of Information} \\
\textit{Florida State University}\\
Tallahassee, FL, USA \\
bb23u@fsu.edu}
\and
\IEEEauthorblockN{Xiao Luo}
\IEEEauthorblockA{\textit{Spears School of Business} \\
\textit{Oklahoma State University}\\
Stillwater, OK, USA \\
xiao.luo@okstate.edu}
\and
\IEEEauthorblockN{Karim Hanna}
\IEEEauthorblockA{\textit{Morsani College of Medicine} \\
\textit{University of South Florida}\\
Tampa, FL, USA \\
khanna@usf.edu}
\and
\IEEEauthorblockN{Mia Liza A. Lustria}
\IEEEauthorblockA{\textit{College of Social Work} \\
\textit{Florida State University}\\
Tallahassee, FL, USA \\
mlustria@fsu.edu}
\and
\IEEEauthorblockN{Carl Yang}
\IEEEauthorblockA{\textit{Department of Computer Science} \\
\textit{Emory University}\\
Atlanta, GA, USA \\
j.carlyang@emory.edu}
\and
\IEEEauthorblockN{Zhe He}\
\IEEEauthorblockA{\textit{School of Information} \\
\textit{Florida State University}\\
Tallahassee, FL, USA \\
zhe@fsu.edu}}
\maketitle

\begin{abstract}
Accurate interpretation of lab results is crucial in clinical medicine, yet most patient portals use universal normal ranges, ignoring conditional factors like age and gender. This study introduces Lab-AI, an interactive system that offers personalized normal ranges using retrieval-augmented generation (RAG) from credible health sources. Lab-AI has two modules: factor retrieval and normal range retrieval. We tested these on 122 lab tests: 40 with conditional factors and 82 without. For tests with factors, normal ranges depend on patient-specific information. Our results show GPT-4-turbo with RAG achieved a 0.948 F1 score for factor retrieval and 0.995 accuracy for normal range retrieval. GPT-4-turbo with RAG outperformed the best non-RAG system by 33.5\% in factor retrieval and showed 132\% and 100\% improvements in question-level and lab-level performance, respectively, for normal range retrieval. These findings highlight Lab-AI's potential to enhance patient understanding of lab results.
\end{abstract}

\begin{IEEEkeywords}
Retrieval Augmented Generation, Lab test Interpretation, Personalized Information Retrieval
\end{IEEEkeywords}

\section{Introduction}
The Health Information Technology for Economic and Clinical Health (HITECH) Act of 2009 played a key role in promoting the adoption and meaningful use of electronic health records (EHRs) throughout the U.S. healthcare system. Through the Medicare and Medicaid EHR Incentive Programs, the Act provided financial incentives that facilitated widespread EHR adoption. In December 2016, the 21st Century Cures Act further advanced healthcare by reinforcing patients' rights to have timely and electronic access to their health information~\cite{lye201821st}. This legislation supported the development and use of patient portals by mandating that patients have electronic access to their health records. Additionally, the Act's anti-information blocking provisions encouraged healthcare organizations to utilize patient portals effectively, ensuring patients can access their health data without unnecessary obstacles. Patient portals are crucial in engaging patients, as they enable individuals to view their medical records, communicate with their healthcare providers, and manage appointments and medications~\cite{dendere2019patient}. 

Even though viewing lab results is the most frequent activity in patient portals, there are also challenges for patients. Lab results are often presented in medical jargon that can be difficult for patients to understand. Lab results typically provide reference ranges (or normal ranges), determined by analyzing lab results from a large, diverse group of healthy individuals~\cite{friedberg2007origin}. The normal range is typically defined as the interval between two values that encompass the middle 95\% of the data (often referred to as the 95\% reference interval). This means that 95\% of healthy individuals' results fall within this range, and 5\% fall outside of it. The normal range is then validated with clinical data and adjusted if necessary. However, lab test results can vary by age, sex, and other factors. For example, the normal range for red blood cell count (RBC) for men is 4.7 to 6.1 million cells/mcL and 4.2 to 5.4 million cells/mcL for women. Unfortunately, mainstream patient portals usually present universal normal ranges for lab tests based on a large population sample as a general baseline without considering the individualized context. Implementing individualized normal ranges requires complex adjustments and may also need input from patients. The emergence of generative large language models (LLMs), like ChatGPT, presents a promising approach to incorporating individual context by prompting patients to provide missing information from their medical records to identify personalized normal ranges. In this work, we explore the use of LLMs with Retrieval-Augmented Generation (RAG) approach to identify the factors that may influence lab test results and to determine individualized normal ranges for lab tests. 

RAG improves the output of a Large Language Model (LLM) by integrating curated information from authoritative external sources, outside the model's original training data, before generating a response~\cite{NEURIPS2020_6b493230}. The external knowledge base is transformed into text embeddings, which are vector representations of natural language that encode its semantic information. Text embeddings play a pivotal role in various natural language processing (NLP) tasks, including information retrieval (IR) and RAG~\cite{zhu2024longembedextendingembeddingmodels}. It is generally reported that RAG, which combines both retrieval and generative techniques, enhances the accuracy and contextual understanding of LLM by incorporating authoritative external sources into their responses. This approach reduces the likelihood of generating incorrect information, offers flexibility to adapt to various domains, and allows for scalability as knowledge bases can be updated without retraining the model. Additionally, RAG improves transparency by listing sources of the response, making it easier to trace the information used, thereby reducing hallucinations, one of the widely reported issues of LLMs.

\begin{figure}
    \centering
    \includegraphics[width=7cm]{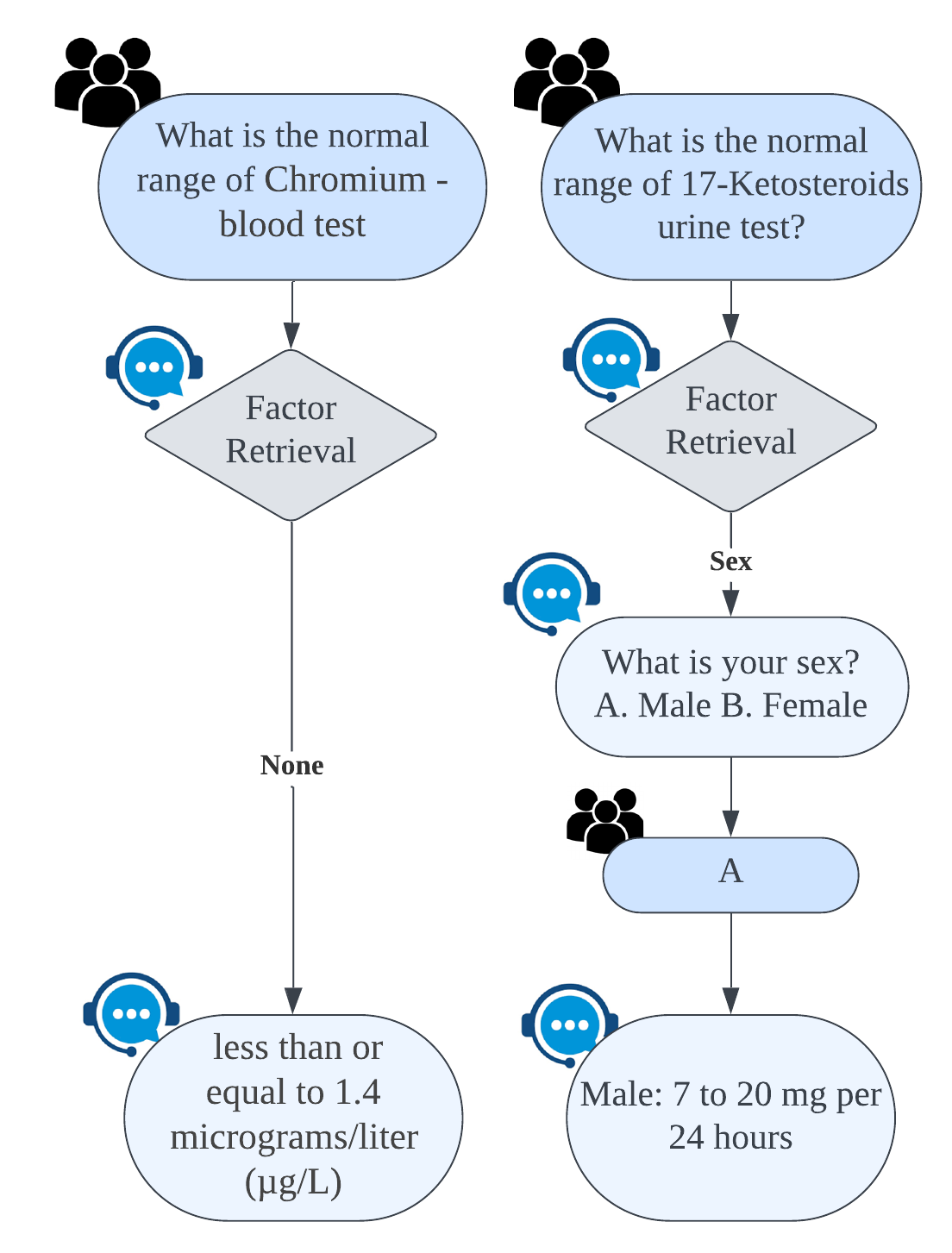}
    \caption{An example interaction with Lab-AI}
    \label{fig:QA}
\end{figure}

 In this study, we present Lab-AI, a framework designed to evaluate the role of LLMs in interpreting lab test normal ranges. Figure 1 illustrates an example interaction with Lab-AI. Lab-AI enables users to interact by inputting questions to produce an individualized normal range of a specific lab test. Upon receiving the query, Lab-AI analyzes the test in question and determines whether additional contextual factors (e.g., gender, age) are required to provide an accurate interpretation of the result. These contextual factors, referred to as "factors", are critical variables that influence normal ranges. If additional factors are required, Lab-AI prompts the user with a set of multiple-choice questions, allowing the user to select the relevant information (e.g., "1. A, 2. B, 3. C"). Based on the user's responses, Lab-AI returns the corresponding individualized normal range. For tests where no additional factors are necessary, the system directly outputs the normal range. The complete Lab-AI system consists of two modules: a factor retrieval module and a normal range retrieval module. Specifically, the primary goal of this research was to explore the feasibility and potential of LLMs in augmenting the interpretation of lab test results, particularly for tests where multiple factors influence normal ranges. This study leverages publicly available, curated, evidence-based lab test information to enhance the understanding of these results. Our findings highlight the potential of generative AI to improve the accessibility and comprehension of lab test data, narrow disparities in access to critical health information, and ultimately contribute to improved patient outcomes. This work paves the way for the responsible integration of AI technologies for lab test interpretation into clinical practice, promoting equitable access to personalized healthcare.

 The main contributions of this study include:

 \begin{itemize}
     \item Development of a dataset to evaluate the proposed system on personalized lab test interpretation.
     \item Construction of RAG-enhanced system to provide more accurate personalized lab test results
     \item Comparison of original chat-bot and RAG-enhanced chat system to show the improvements
 \end{itemize}

\section{Related Work}
RAG has been successfully applied in clinical applications such as interpreting pharmacogenomics lab testing results \cite{Murugan2024}, improving patient admission prediction \cite{Glicksberg2024}, summarizing and extracting malnutrition information \cite{Alkhalaf2024}, and for treatment recommendations and medical guidelines \cite{Zakka2024}. Murugan et al. evaluated a GPT-4-based AI assistant with RAG to enhance pharmacogenomic decision-making by interpreting PGx test results, using data from the Clinical Pharmacogenetics Implementation Consortium~\cite{Murugan2024}. The study demonstrated high efficacy in addressing specialized provider queries and emphasized the potential for improving accuracy, relevance, and ethical considerations for wider implementation of LLM with RAG in clinical practice. Alkhalaf et al. tested zero-shot prompt engineering in LLaMa 2, both alone and with RAG, for summarizing and extracting malnutrition data from EHRs in aged care facilities~\cite{Alkhalaf2024}. LLaMa 2 achieved high accuracy in summarizing nutritional status (93.25\%) and extracting risk factors (90\%). Implementing RAG improved summarization accuracy to 99.25\% and reduced hallucinations but it did not further enhance risk factor extraction. Zakka et al. evaluated the performance of a RAG model called Almanac for clinical decision-making, and compared it with standard large language models like GPT-4, Bing, and Bard~\cite{Zakka2024}. RAG was implemented by integrating curated medical resources such as PubMed, UpToDate, and BMJ Best Practices with the model to enhance factuality, completeness, and adversarial safety in response to clinical questions. The study reported that Almanac significantly outperformed other models in factuality and completeness, particularly in adversarial settings, demonstrating the potential for retrieval-augmented models to improve clinical decision support.

LLMs are increasingly being used for interpreting lab test results in clinical contexts, offering potential enhancements in accuracy and efficiency. He et al. conducted a study comparing the quality of responses from various LLMs, including GPT-4, GPT-3.5, LLaMA 2, MedAlpaca, and ORCA mini, to questions related to lab test results on a social Q\&A platform Yahoo! Answers~\cite{He2024}. They found that GPT-4 outperformed other models, achieving higher scores in accuracy, helpfulness, relevance, and safety, with superior handling of complex lab tests such as hemoglobin A1c, bilirubin, and lipid panels. Munoz-Zuluaga et al. evaluated the ability of GPT-4 to answer a representative set of questions frequently encountered in the laboratory medicine field, ranging from basic knowledge to complex interpretation of laboratory data in a clinical context~\cite{Munoz-Zuluaga2023}. They found that while GPT-4 correctly answered 50.7\% of the questions, it also produced incomplete, incorrect, or irrelevant responses in the remaining cases. Cadamuro et al. reported that while GPT-4 could recognize and provide basic interpretations of individual lab tests such as glucose, HbA1c, and complete blood count (CBC), it struggled with integrating these results into a coherent clinical interpretation, often failing to recognize conditions like Gilbert's syndrome or pre-analytical issues such as non-fasting glucose samples  ~\cite{Cadamuro2023}. Meyer et al. compared the performance of GPT-4, Gemini, and Le Chat in interpreting laboratory questions related to CBC from the online health forum ‘AskDocs’ subreddit on Reddit~\cite{Meyer2024}. The study found that while GPT-4 and the other chatbots provided empathetic and structured responses, their accuracy and medical reliability were significantly lower than that of online physicians. 

These studies underscore the critical role of LLMs in interpreting lab test results, highlighting both their potential and current limitations. While LLMs like GPT-4 have demonstrated considerable accuracy in certain scenarios, such as interpreting complex cases involving serum lipase and blood lead levels, they still fall short compared to human experts in ensuring consistent and reliable results across a broad spectrum of laboratory data. Moreover, none of these prior studies evaluated LLMs' capability in determining the normalcy of lab results by considering individual information. The findings emphasize the necessity for continued development and rigorous validation of LLMs to enhance their utility in clinical settings, where precise and accurate interpretation of lab tests is critical for patient safety. The current study employs RAG with LLMs for interpreting lab test data, especially the normal ranges. By integrating real-time retrieval of lab test-relevant information through RAG, the study aims to enhance the accuracy and reliability of LLMs in interpreting lab test data, thereby improving clinical decision-making processes.

\section{Methods}
\subsection{System Overview}
In this study, we utilized a RAG-enhanced LLM system, Lab-AI, to help the patients find an authoritative lab normal range. Our objective is to design an AI assistant that could incorporate knowledge from credible sources and provide an accurate normal range by considering individual information, leveraging the power of the LLMs. We employed a RAG-Sequence model, which retrieves documents and generates a full response based on each document individually. For each retrieved document, the model generates a complete response independently. It then selects the best response by comparing the likelihood scores of the different generated sequences\cite{NEURIPS2020_6b493230}. With the support of RAG, which synergistically merges LLMs’ intrinsic knowledge with the vast, dynamic repositories of external databases~\cite{gao2024retrievalaugmentedgenerationlargelanguage}, the accuracy and credibility of the
generation will be enhanced.

\begin{figure*}
    \centering
    \includegraphics[width=12cm]{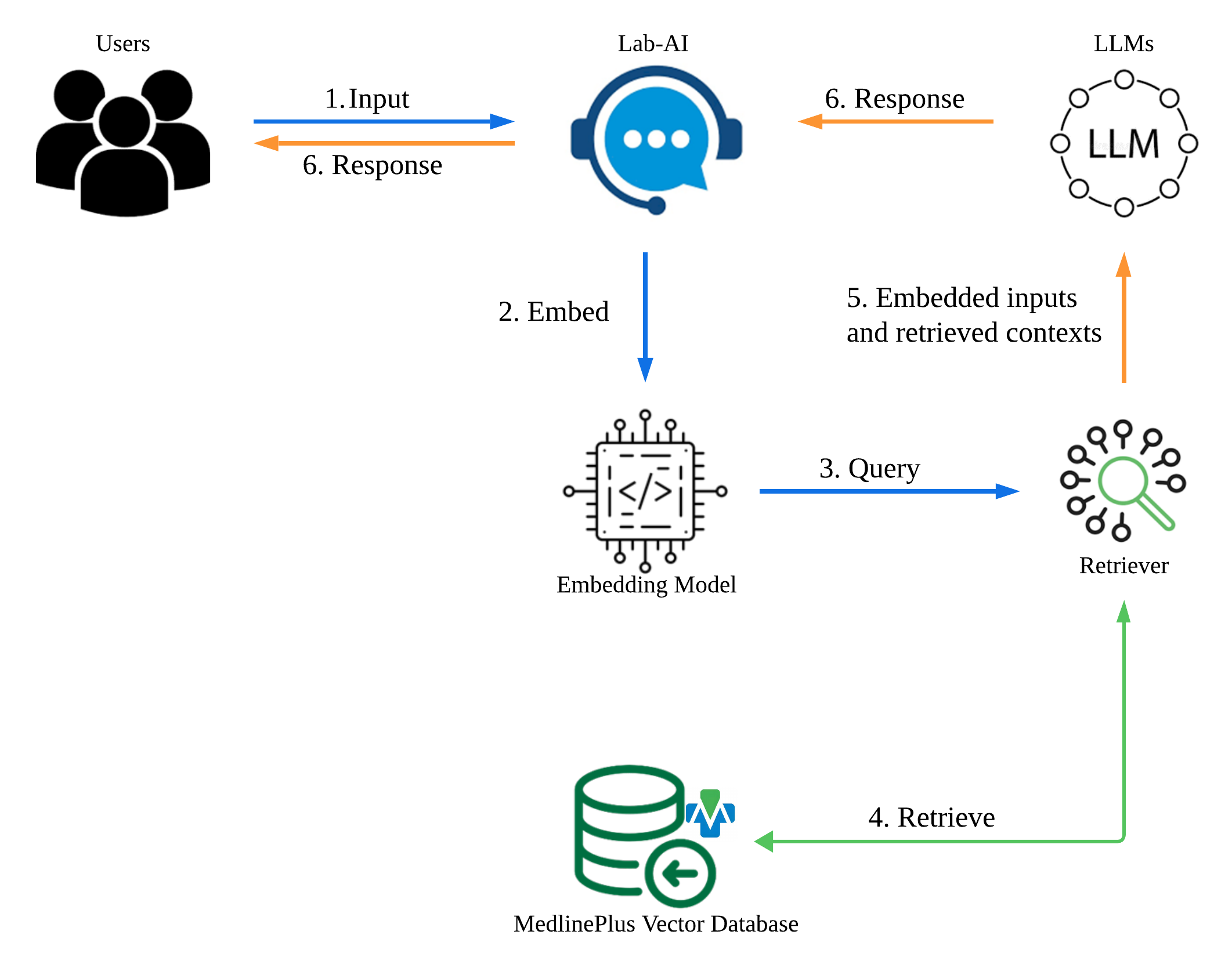}
    \caption{Interactions with Lab-AI}
    \label{fig:workflow}
\end{figure*}

Figure \ref{fig:workflow} illustrates users' interactions with Lab-AI. The tool will start with a user's question. The format of questions can vary but we only tested questions like "What is the normal range for [lab test name]?". For each user input, the system employs OpenAI’s "text-embedding-3-large" model to embed the query. The embedding model is designed to act as a text encoder, transforming the input into a high-dimensional vector representation. The dimensions of these embedded queries are consistent with the data stored in the vector database, allowing for efficient and accurate retrieval of information.

 The embedded question will then be sent to the retriever as a query. The retriever will compare the embedded input against the data stored in the vector database and return the relevant matches. Using similarity metrics, the system retrieves from the vector database the top two documents that have the highest similarity scores to the query and returns them to the retriever. To connect the retriever and LLMs, we used "chat\_engine" which is a stateful analogy of a query engine to keep track of the conversation history from Llama-index~\cite{Liu_LlamaIndex_2022}. The embedded inputs and retrieved contents will then be sent to the LLMs. LLMs mainly refer to transformer-based neural language models that contain tens to hundreds of billions of parameters, which are pre-trained on massive text data and exhibit strong language understanding and generation abilities~\cite{minaee2024largelanguagemodelssurvey}. Based on the provided context, LLMs generate responses by processing the embedded inputs, retrieved contexts and system prompts. These responses are then redirected to the user through the Lab-AI system. The following section outlines the process of constructing and evaluating two important modules in Lab-AI, namely, factor retrieval module and normal range retrieval module.

\subsection{Step 1: Vector Database Generation} 
\begin{figure*}[htbp]
    \centering
    \includegraphics[width=18cm]{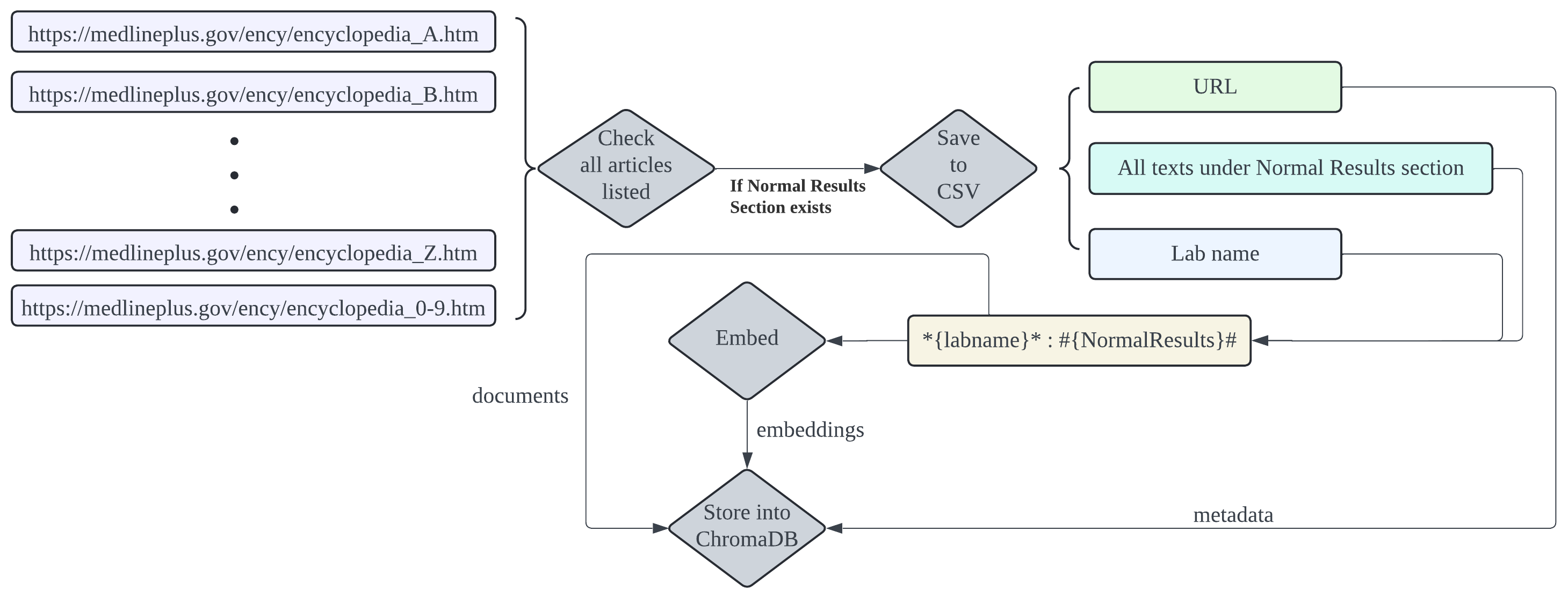}
    \caption{Data Crawl and Vector Database Generation}
    \label{fig:vector database}
\end{figure*}

We obtained a set of lab test articles from MedlinePlus~\cite{medlineplus}, a credible online health information resource developed and maintained by the U.S. National Library of Medicine. Following the pipeline depicted in Figure \ref{fig:vector database}, we crawled data from the medical encyclopedia pages by iterating over the initials from A to Z and 0-9. For each page, we specifically examined the "Normal Results" section. If the web document included this section, we stored the lab name, the corresponding text under "Normal Results" and the URL. To facilitate the extraction of lab test information from these articles, we used Beautiful Soup, a Python package designed for parsing HTML and XML markup documents~\cite{richardson2007beautiful}. The vector database is an efficient vector data storage that allows fast and accurate similarity search and retrieval~\cite{han2023comprehensivesurveyvectordatabase}. We first reformatted the data into the format "[labname]: {NormalResult}". We opted to use "text-embedding-3-large" from OpenAI as our embedding model and converted all lab normal ranges into 3072-dimensional vectors. Additionally, we included URLs as metadata so that the RAG system can link to the appropriate source on MedlinePlus when necessary. We chose ChromaDB~\cite{ChromaDB} as the vector database due to its compatibility with OpenAI embedding functions.

\subsection{Step 2: Data Preparation}

\begin{figure}
\centering
    \includegraphics[width=0.9\linewidth]{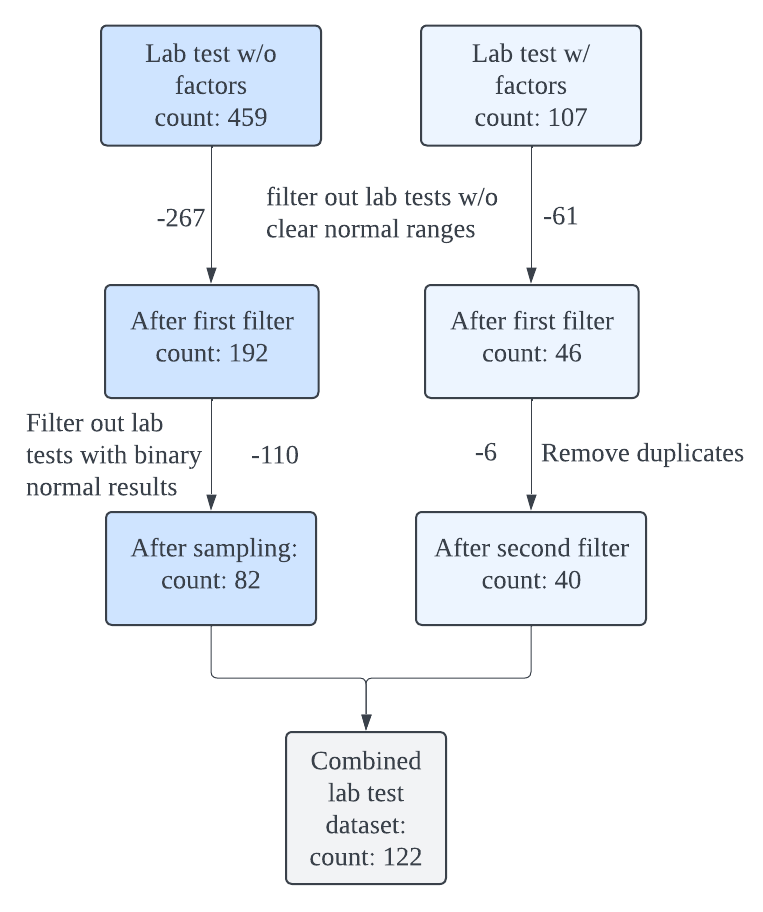}
    \caption{Lab test selection}
    \label{fig:lab test selection}
\end{figure}

As shown in Figure \ref{fig:lab test selection}, a total of 566 lab tests were identified, 107 of which had associated factors or consisted of multiple tests, while the remaining 459 did not have factors. We manually filtered out the tests with factors that lacked normal ranges, as well as dropped lab panels which are comprised of multiple tests as duplicates. After filtering, the final number of lab tests with factors was 40. A similar filtering process was applied to the tests without factors, reducing their number from 459 to 192. From these, we further filtered out lab tests with binary normal results, for example, the normal range is "negative" while the abnormal range is "positive". Our focus was primarily on lab tests using blood or urine as specimens. Over 62\% of the tests in the convenience sample were blood tests, while 21\% used urine. Other specimens included tissues, fluids, marrow, and others.

Our study consists of two experiments. The first experiment was designed to test the ability of RAG-enhanced LLM systems to retrieve the factors associated with lab tests. The second experiment evaluates whether RAG-enhanced LLM systems can accurately retrieve the correct normal ranges given different factors. Non-RAG LLMs serve as the baseline for comparison. Consequently, two datasets were created for these experiments.

\subsubsection{Lab Test Factor Dataset}
% Lab result interpretation dataset
This dataset supports our first objective — to evaluate if RAG-enhanced LLM systems can accurately extract and prompt the key factors determining the normal range for a lab test. To create the ground truth, we first manually extracted the factors for determining normal reference ranges for the included 122 lab tests. Table \ref{Factor data sample} provides example factors for lab tests. We only considered the factors that determine different normal ranges. Factors that may affect the lab test but do not have a specific normal range in the article were excluded. For example, the normal range of "Aldolase" from MedlinePlus is: "Normal results range between 1.0 to 7.5 units per liter (0.02 to 0.13 microkat/L). There is a slight difference between men and women." Although the normal range for men and women can be slightly different, we did not count gender as a factor because they do not have specific normal ranges in the article.

\begin{table*}[h]
    \centering
    \vspace{0.2cm}
    \caption{Example factors for lab test normal ranges}
    \vspace{-0.2cm}
    \label{Factor data sample}
    
    \begin{tabular}{|l|l|}
    \hline
        Lab Test                                   & Factors                         \\ \hline
        Aldolase blood test                        & None                            \\
        Prostate-specific antigen (PSA) blood test & Age                             \\
        Erythrocyte sedimentation rate (ESR)       & Age, Sex                        \\
        Luteinizing hormone   (LH) blood test      & Age, Sex, Menstrual cycle phase \\
        \hline
    \end{tabular}
\end{table*}

\begin{figure}

    \centering
     \includegraphics[width=8cm]{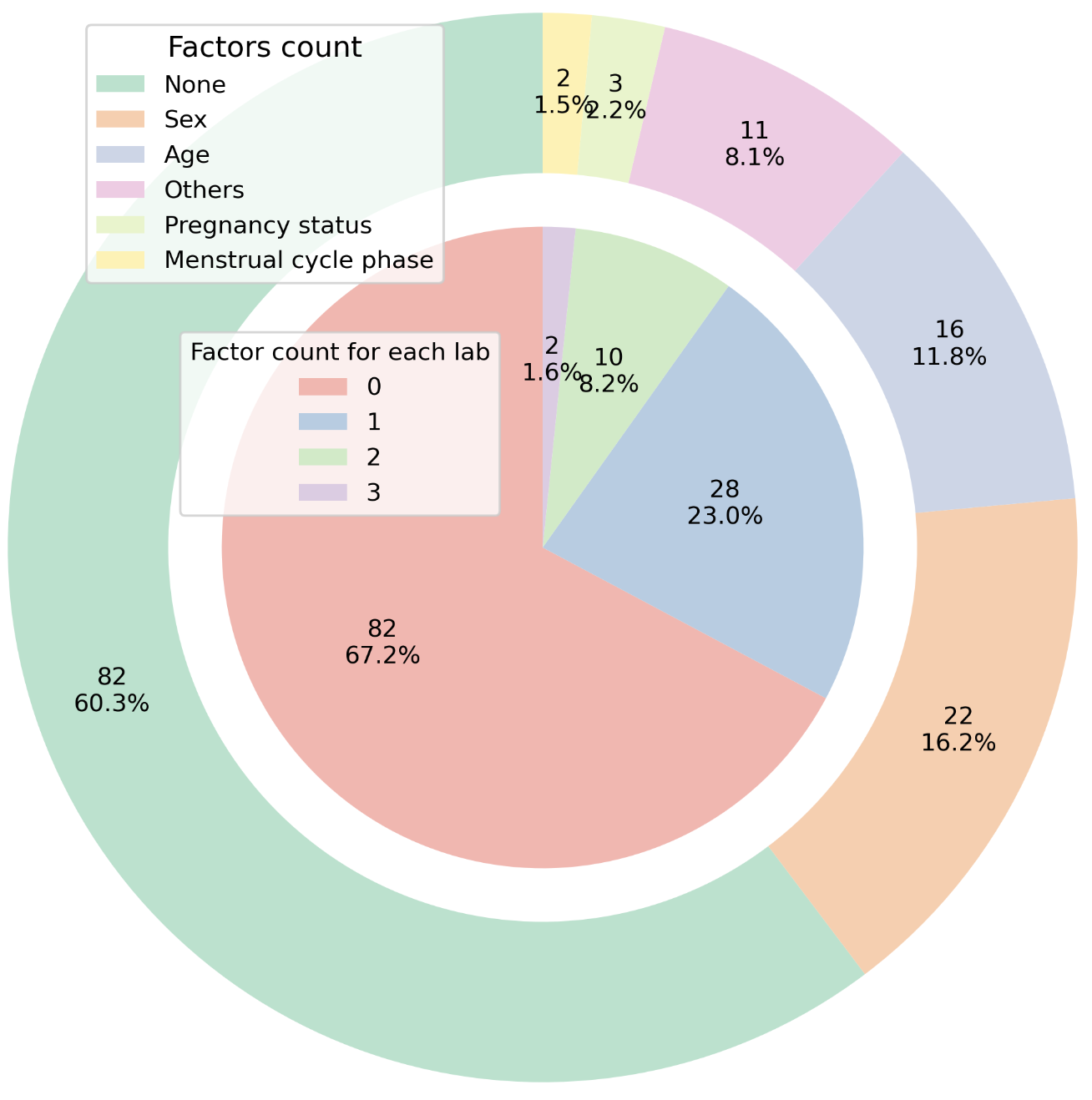}
     \vspace{0.5cm}
    \caption{Factor Dataset Statistics}
    \label{piegraph}
\end{figure}

After manual review, we found that all the lab tests had up to three factors influencing their normal ranges, such as age, gender, or specimen type. Figure \ref{piegraph} shows the dataset distribution. The donut chart illustrates the frequency of each factor, while the pie chart displays the distribution of how many factors each lab test includes. Notably, 67.2\% of the selected lab tests had no associated factors, indicating a universal normal range. Among the remaining tests, 22 depended on sex, 16 on age, 3 on pregnancy status, and 2 were influenced by specific conditions related to women. The others include uncommon cases, such as "Is the patient an athlete?" or "Is the patient standing or lying down?"

\subsubsection{Lab Test Normal Range Dataset}
This dataset was developed to determine whether RAG-enhanced LLM systems can provide correct normal ranges given specific factors. For a lab test with multiple factors, we generated questions with all the different combinations of these factors. For example, the erythrocyte sedimentation rate (ESR) test has two determining factors: gender and age. We generated four question prompts based on the normal reference range: 1) "What is the normal range of the erythrocyte sedimentation rate (ESR) test for females over 50?" 2) "What is the normal range of the erythrocyte sedimentation rate (ESR) test for females under 50?" 3) "What is the normal range of the erythrocyte sedimentation rate (ESR) test for males over 50?" 4) "What is the normal range of the erythrocyte sedimentation rate (ESR) test for males under 50?" For lab tests without factors, the question prompt was simply: "What is the normal range of [lab test name]?" A total of 212 questions were generated from 122 lab tests. The distribution was as follows: 82 questions were derived from 82 individual lab tests, 55 questions from 28 lab tests, 65 questions from 10 lab tests, and 10 questions from 2 lab tests.

\subsection{Step 3: Experiments of factor retrieval and normal range retrieval}
Factor retrieval is a crucial step in identifying the key factors determining the accurate normal range for a specific lab test. We designed a universal prompt and tested it across three OpenAI LLMs: GPT-3.5-turbo, GPT-4-turbo, and GPT-4 to assess if these LLMs can identify the critical factors associated with the normal ranges of a lab test. 

We maintained the default settings, except for setting the temperature to zero to ensure consistency and minimize randomness. The prompt used for LLMs can significantly influence the quality of their responses. Initially, we crafted a simple prompt asking the LLMs which factors might influence the normal range for a given lab test, expecting a list of factors in return. However, the LLMs returned detailed factors and their corresponding normal ranges, and the RAG system failed to retrieve information from our stored vector database as expected. To refine the prompt, we consulted GPT-4, which helped us improve its structure. The final prompt used a step-by-step instruction format: first, analyze the information from the vector database, then inform the LLM that we are only interested in factors associated with a numeric value. The LLM was instructed to output only those key factors and respond with "None" if no factors were relevant. For GPTs without RAG, we omitted the first step since it does not require information retrieval from the vector database. Additionally, we used a 2-shot example to help the LLMs better understand the desired output.

\begin{figure*}[htbp]
    \centering\includegraphics[width=\linewidth]{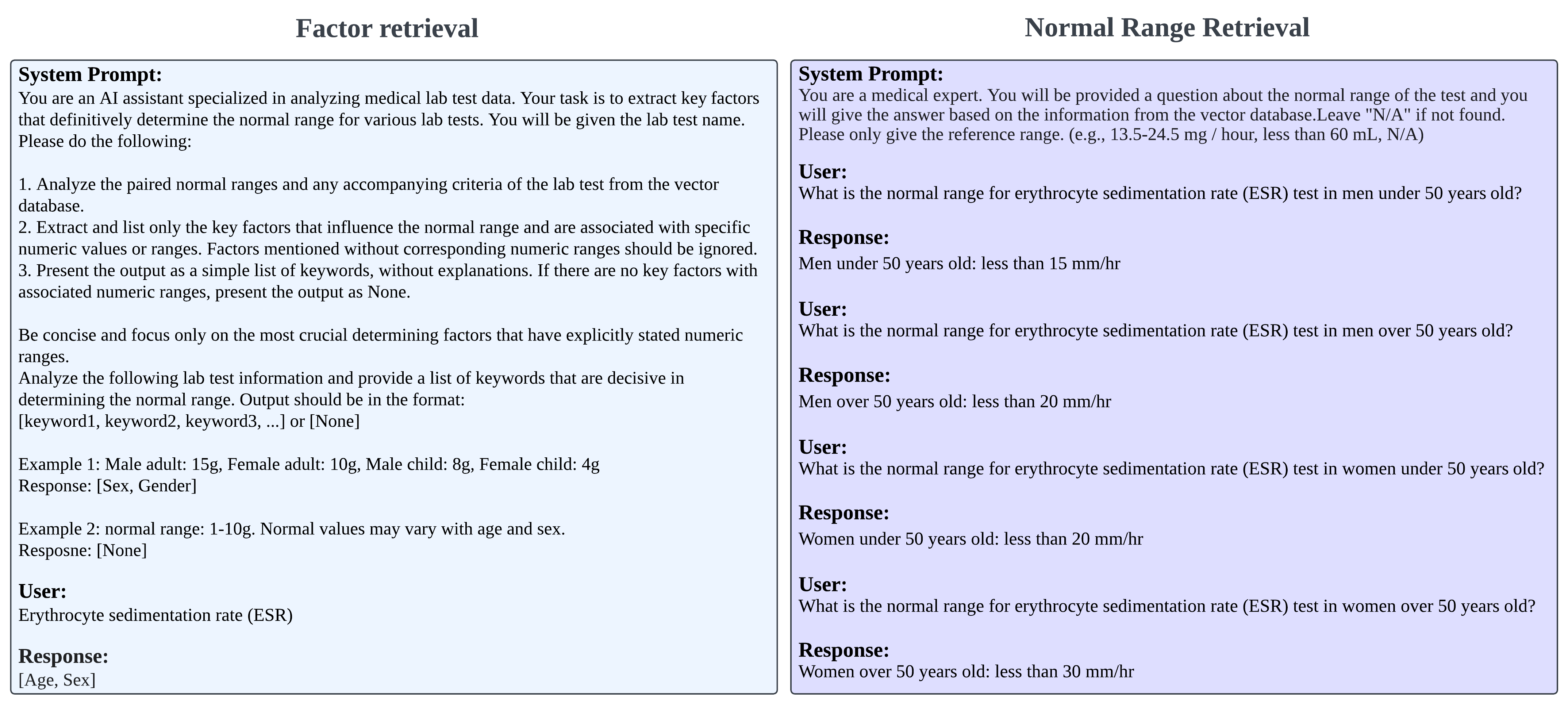}
    \caption{Prompts used for the two modules of Lab-AI}
    \label{prompt and response}
\end{figure*}

For evaluation, we summed up all of the True Positives, False Positives, and False Negatives first, then calculated the corresponding precision, recall, and F-1 score for both RAG-based and non-RAG-based systems based on three selected LLMs by Equation \ref{eq:prec}, \ref{eq:rec} and \ref{eq:f1}, where \(TP_i\), \(FP_i\), \(FN_i\) are the count of True Positives, False Negatives and False Negatives of \(i^{th}\) lab. Figure \ref{prompt and response} shows the prompts used in the experiments.

\begin{align}
precision &= \frac{\sum_{i=1}^{N} TP_i  }{\sum_{i=1}^{N} (TP_i + FP_i)}\label{eq:prec}
\\\nonumber\\
recall &= 
\frac{\sum_{i=1}^{N} TP_i  }{\sum_{i=1}^{N} (TP_i + FN_i)}\label{eq:rec}
\\\nonumber\\
F1 &= 2 \times \frac{precision \times recall}{precision + recall} \label{eq:f1}
\end{align}

Accurate normal range retrieval is the primary goal of our study. We selected the best-performing model from the factor retrieval task, as accurately detecting lab test factors is the prerequisite for retrieving the correct normal range. Similar to the first task, we minimized response randomness by setting the temperature to zero. The system prompt was generated through iterative refinement using GPT-4-turbo, incorporating some examples of desired outputs. The user prompt is a straightforward question requesting the normal range given specific factor values, e.g., "What is the normal range for urine 17-ketosteroids test in males?" 

The expected response is a precise value or range with appropriate units. However, it is unfair to directly compare the retrieved values between LLMs and RAG-enhanced LLM systems since we already let the RAG system know the expected values. As a result, when we evaluated the non-RAG LLMs' results, we used many other credible lab test references, including MKSAP~\cite{american2021mksap}, LOINC~\cite{loinc-302}, IAPAC~\cite{IAPAC_Lab_Values}, American Board of Internal Medicine~\cite{ABIM_Lab_Values}, Health Encyclopedia by University of Rochester Medical Center~\cite{URMC_Lab_Values} and ARUP laboratories~\cite{ARUP_Lab_Testing}. If non-RAG LLMs' response matches any of those references, we considered that response as correct. For the RAG system, only the exact match was considered correct. This evaluation could demonstrate the superiority of the RAG-based systems. Both question-level and lab-level accuracy will be used as metrics for this task. Question-level accuracy (QLA) is defined as the ratio of the number of questions with correctly retrieved ranges to the total number of questions or mathematically computed by Equation \ref{eq:qla}, where T is the total number of questions generated. Lab-level accuracy(LLA) is calculated as the sum of the average accuracy values of questions for each lab test over all the tests divided by the total number of lab tests or computed by Equation \ref{eq:lla}, where question\(_{i,j}\) is the \(j^{th}\) question of the \(i^{th} lab\).

\begin{align}
QLA &= \frac{1}{T} \sum_{i=1}^{T} \mathbb{I}\{\text{answer to question}_{i} \text{ is correct}\} \nonumber
\\ 
&\textit{\qquad\qquad\qquad\qquad where } T = \sum_{i=1}^{N}\sum_{j=1}^{n_i}\mathbf{1}\label{eq:qla}
\\\nonumber\\
LLA &= \frac{1}{N} \sum_{i=1}^{N} \frac{\sum_{j=1}^{n_i}\mathbb{I}\{\text{answer to question}_{i,j} \text{ is correct}\}}{n_i}\label{eq:lla}
\end{align}

% where question i,j means the j^{th}\text{ question of the }i^{th}\text{ lab}

\section{Results}

\subsection{Lab Test Factor Retrieval}
For the factor retrieval task, GPT-4-turbo outperformed the other models in both RAG and non-RAG systems. Surprisingly, without the RAG system, GPT-3.5-turbo showed better results than GPT-4, achieving significantly higher precision and F1, although it did not perform as well with RAG, as seen in Table \ref{factor retrieve performance} where GPT-4-turbo with RAG achieved an F1 score of 0.948. Table \ref{table: factor error case} lists some cases of the RAG-based system with different GPT models. GPT-3.5-turbo with RAG will respond with choice-level factors. We cannot say it is completely wrong, but it did not follow our system prompt requirement. GPT-4 with RAG tended to provide extra factors in addition to the true labels. Some of those factors were mentioned in MedlinePlus but are not closely associated with a clear normal range. Some factors may have come from its own knowledge base. GPT-4-turbo with RAG showed a more balanced performance, but it may also prompt additional factors or miss some factors.

\subsection{Lab Test Normal Range Retrieval}
Since GPT-4-turbo with RAG outperformed the other models, we further evaluated it in the normal range retrieval task. Table \ref{table: normal range performance} shows the performance of normal range retrieval at the question and lab levels. GPT-4-turbo without RAG performed poorly with only an accuracy of 42.9\%  at the question level and 49.6\% at the lab level. GPT-4-turbo with RAG performed exceptionally well, with only one error out of 212 generated questions. For lab tests with relevant factors, the accuracy reached 100\%. Even though we also compared the results from the non-RAG GPT-4-turbo with multiple credible sources, its performance still remained unsatisfactory. Table \ref{table: normal range error case} highlights the single error, along with a similar case in which GPT-4-turbo with RAG provided a correct response. Unlike most lab tests, the normal ranges for the acid-fast stain and anti-smooth muscle antibody tests are "no acid-fast bacteria found" and "no antibodies found," respectively. Although these results could be interpreted as zero, we did not apply any preprocessing for these cases.
% \section*{Discussion}
% In this study, our aim was to develop an interactive Lab-AI system capable of providing accurate normal reference ranges for lab tests specific to each patient. This is achieved by first identifying the factors that affect normal reference ranges, and then retrieving the appropriate range based on the patient's responses to those factors. According to the results in Table \ref{factor retrieve performance}, GPT-4-turbo 
% with RAG achieved an F1 of 0.95 on this task, whereas the original GPT-4-turbo reached only an F1 of 0.659. Typically, GPT-4, with its larger number of parameters compared to GPT-3.5-turbo, would be expected to outperform GPT-3.5-turbo. However, in the factor retrieval task, GPT-4 yielded lower precision and F1. While GPT-4-turbo achieved the highest recall, making it the most effective non-RAG model at retrieving relevant factors, its recall was still below 0.7. With RAG, GPT-4-turbo outperformed other models in retrieving accurate and relevant cases. GPT-4 with RAG significantly outperformed GPT-3.5-turbo with RAG, indicating better compatibility between GPT-4 and the RAG system. Interestingly, while GPT-4 with RAG excelled at retrieving relevant factors, it was also more prone to providing factors not listed in the original source. This may be due to its tendency for hallucinations, whereas GPT-4-turbo is more stable and less prone to generating creative but incorrect information. 

\begin{table*}[htbp]
\centering
\vspace{0.5cm}
\caption{Factor Retrieval Performance}
% \vspace{-0.2cm}
\scalebox{1.2}{
\begin{tabular}{c|ccc|ccc}
\multicolumn{1}{l|}{} & \multicolumn{3}{c|}{Non-RAG} & \multicolumn{3}{c}{RAG}    \\ \hline
                      & Precision  & Recall  & F-1    & Precision & Recall & F-1   \\
GPT-3.5-turbo         & 0.65      & 0.679   & 0.664  & 0.230     & 0.478  & 0.311 \\
GPT-4-turbo &
  \cellcolor[HTML]{DAE8FC}0.690 &
  \cellcolor[HTML]{DAE8FC}0.731 &
  \cellcolor[HTML]{DAE8FC}0.710 &
  \cellcolor[HTML]{DAE8FC}0.948 &
  \cellcolor[HTML]{DAE8FC}0.948 &
  \cellcolor[HTML]{DAE8FC}0.948 \\
GPT-4                 & 0.529      & 0.672   & 0.592  & 0.747     & 0.881  & 0.808
\end{tabular}
}
\label{factor retrieve performance}
\end{table*}

% [Zhe: You should add a table with example cases in which LLMs did not prompt all the conditions. ]

\begin{table*}[h!]
\centering
\renewcommand{\arraystretch}{1.5} % Adjust row height

\caption{Sample Error Cases in Factor Retrieval Evaluation}
\vspace{0cm}
\label{table: factor error case}
\scalebox{1.1}{\begin{tabular}{|>{\centering\arraybackslash}m{3cm}|>{\centering\arraybackslash}m{3cm}|>{\centering\arraybackslash}m{4cm}|>{\centering\arraybackslash}m{5cm}|}
\hline
Model & Lab Test & Ground Truth Factors & Model Response \\ \hline
\multirow{2}{3cm}[-1ex]{\centering GPT-3.5-turbo} & BUN - blood test & None & Age, Renal function \\ 
\cdashline{2-4}
 & HCG blood test - quantitative & Pregnancy status, Sex & Quantitative \\ \hline
\multirow{2}{3cm}[-1ex]{\centering GPT-4} & Ammonia blood test & None & Age, Fasting \\ 
\cdashline{2-4}
 & HCG blood test - quantitative & Pregnancy status, Sex & None \\ \hline
\multirow{2}{3cm}[-1ex]{\centering GPT-4-turbo} & Creatinine urine test & Sex & None \\
 \cdashline{2-4}
 & Serum progesterone & Sex, Menstrual cycle phase, Pregnancy status & Sex, Menstrual cycle phase \\
\hline

\multirow{2}{3cm}[-2ex]{\centering GPT-3.5-turbo with RAG} & DHEA-sulfate test & Sex, Age & Sex, Age, Female, Male \\ 
\cdashline{2-4}
 & Estradiol blood test & Sex, Age & Sex, Age, Male, Female, Premenopausal, Postmenopausal, pg/mL, pmol/L \\ \hline
\multirow{2}{3cm}[-1ex]{\centering GPT-4 with RAG} & 17-OH progesterone & Age & Age, Birth weight \\ 
\cdashline{2-4}
 & Acid loading test (pH) & None & Urine pH \\ \hline
\multirow{3}{3cm}[-2ex]{\centering GPT-4-turbo with RAG} & Creatinine urine test & Sex & Sex, Age \\
 \cdashline{2-4}
 & Serum progesterone & Sex, Menstrual cycle phase, Pregnancy status & Menstrual cycle phase, Pregnancy status \\
\cdashline{2-4}
 & Urine concentration test & Water consumption & None \\ 
 \hline
\end{tabular}
}
\end{table*}

\begin{table*}[h!]
\centering
\vspace{0.25cm}
\caption{Normal Range Retrieval Performance}
\scalebox{1.2}{
\begin{tabular}{|cc|ccc|ccc|}
\hline
\multicolumn{2}{|c|}{\multirow{2}{*}{}} & \multicolumn{3}{c|}{GPT-4-turbo} & \multicolumn{3}{c|}{GPT-4-turbo with RAG} \\
\multicolumn{2}{|c|}{} & w/ factor & w/o factor & Overall & w/ factor & w/o factor & Overall \\ \hline
\multicolumn{1}{|c|}{\multirow{3}{*}{Question Level}} & \# of total & 130 & 82 & 212 & 130 & 82 & 212 \\
\multicolumn{1}{|c|}{} & \# of correct & 47 & 44 & 91 & 130 & 81 & 211 \\
\multicolumn{1}{|c|}{} & Accuracy & 0.362 & 0.537 & 0.429 & 1.0 & 0.988 & 0.995 \\ \hline
\multicolumn{1}{|l|}{\multirow{3}{*}{Lab Level}} & \# of total & 40 & 82 & 122 & 40 & 82 & 122 \\
\multicolumn{1}{|l|}{} & \# of correct & 16.5 & 44 & 60.5 & 40 & 81 & 121 \\
\multicolumn{1}{|l|}{} & Accuracy & 0.413 & 0.537 & 0.496 & 1.0 & 0.988 & 0.992 \\ \hline
\end{tabular}
}
% \caption{Normal Range Retrieval Performance}
\label{table: normal range performance}
\end{table*}

\begin{table*}[h!]
\centering
\renewcommand{\arraystretch}{1.5} % Adjust the row height
\vspace{0.25cm}
\caption{Normal Range Retrieval Sample Cases using RAG}
\label{table: normal range error case}
\scalebox{1.1}{
\begin{tabular}{|>{\centering\arraybackslash}m{4cm}|>{\centering\arraybackslash}m{6cm}|>{\centering\arraybackslash}m{5.5cm}|}
\hline
Question & Ground Truth Response & GPT-4-turbo with RAG \\ \hline
What is the normal range for Acid-fast stain? & A normal result means no acid-fast bacteria were found on the stained sample & N/A \\ \hline
What is the normal range for Anti-smooth muscle antibody? & Normally, there are no antibodies present & Normally, there are no antibodies present. \\ \hline
\end{tabular}
}
\end{table*}

\section{Discussion}
In this study, our aim was to develop an interactive Lab-AI system capable of providing accurate individualized normal reference ranges for lab tests specific to each patient. This is achieved by first identifying the factors that affect normal reference ranges, and then retrieving the appropriate range based on the patient's responses to those personalized factors. According to the results in Table \ref{factor retrieve performance}, GPT-4-turbo 
with RAG achieved an F1 of 0.948 on this task, whereas the original GPT-4-turbo reached only an F1 of 0.710. Typically, GPT-4, with its larger number of parameters compared to GPT-3.5-turbo, would be expected to outperform GPT-3.5-turbo. However, in the factor retrieval task, GPT-4 yielded lower precision and F1. While GPT-4-turbo achieved the highest recall, making it the most effective non-RAG model at retrieving relevant factors, its recall was still below 0.75. With RAG, GPT-4-turbo outperformed other models in retrieving accurate and relevant cases. GPT-4 with RAG significantly outperformed GPT-3.5-turbo with RAG, indicating better compatibility between GPT-4 and the RAG system. Interestingly, while GPT-4 with RAG excelled at retrieving relevant factors, it was also more prone to providing factors not listed in the original source. This may be due to its tendency for hallucinations, whereas GPT-4-turbo is more stable and less prone to generating creative but incorrect information. 

We further evaluated the RAG system's ability to retrieve normal reference ranges using the best factor retrieval model. Using GPT-4-turbo as the baseline, we relaxed the correctness criterion from an exact match with MedlinePlus to a match with any credible source. While GPT-4-turbo achieved less than 50\% accuracy, GPT-4-turbo with RAG correctly retrieved nearly all normal ranges, making only one error. We implemented two evaluation methods: question-level accuracy, which measures the accuracy across all 212 questions, and lab-level accuracy, which assesses performance across 122 lab tests. The overall accuracy of the RAG system at the question level was 0.995, approximately 132\% higher than the baseline model, while the lab-level accuracy was 0.992, about 100\% higher. According to Table \ref{table: normal range error case}, the only mistake GPT-4-turbo with RAG made was for the acid-fast stain test, where the correct result should have been "no acid-fast bacteria found," but the system responded with "N/A" instead. We also examined a similar case involving the anti-smooth muscle antibody test, where GPT-4-turbo with RAG successfully indicated the correct normal result of "no antibodies present". These cases are challenging because their normal results are non-numeric.

In future studies, we plan to explore the latest LLMs, including GPT-4o, Claude 3.5, and Llama 3.1 in both tasks. Improving the accuracy of factor retrieval could improve the normal range retrieval process further. Building on the success of factor and normal range retrieval demonstrated with the RAG system, we will also focus on designing a user interface that enables patients to interact directly with Lab-AI. We will evaluate Lab-AI with other question prompts such as "What does my result of creatinine urine test of 250 µmol/kg/day mean?" and see if Lab-AI can prompt the patient to provide necessary factor necessary (e.g., gender in this case). Furthermore, relying solely on MedlinePlus as a reference source has limitations. While MedlinePlus is an authoritative resource, laboratory reference ranges can vary across different facilities and testing methods. In future work, we plan to incorporate multiple reliable reference sources and implement a feature that allows patients to specify their laboratory's standard reference ranges, enabling more personalized and accurate result interpretations.

\section{Conclusion}
In this proof-of-concept study, we demonstrated that advanced large language models (LLMs) integrated with Retrieval-Augmented Generation (RAG) can accurately retrieve personalized normal reference ranges for lab tests, significantly outperforming non-RAG models in both factor retrieval and normal range retrieval tasks. The low accuracy of non-RAG GPT models in retrieving normal ranges underscores the unreliability of using GPT alone for this purpose. In contrast, the RAG system effectively connects LLMs to credible health sources, reducing hallucinations and improving reliability. Given the promising results, with an F-1 score of 0.948 for factor retrieval and 0.995 accuracy for normal range retrieval, we are now advancing to the next stage of developing Lab-AI further. This will include generating tailored questions and integrating it into clinical practice to enhance patients' understanding of lab results, shared decision making, and ultimately outcomes.

\section*{Acknowledgment}

This work was supported by the Agency for Healthcare Research and Quality grant R21HS029969 (PI: Z.H.). This project was also partially supported by the University of Florida-Florida State University Clinical and Translational Science Award, which is supported in part by the National Institutes of Health (NIH) National Center for Advancing Translational Sciences under award UL1TR001427.

\bibliographystyle{IEEEtran}
\bibliography{mybibliography}

\end{document}